\documentclass[conference]{IEEEtran}
\IEEEoverridecommandlockouts

\usepackage{cite}
\usepackage{amsmath,amssymb,amsfonts}
\usepackage{algorithmic}
\usepackage{graphicx}
\usepackage{textcomp}
\usepackage{xcolor}

\usepackage{multirow} 
\usepackage{booktabs}
\usepackage{graphicx} 
\usepackage{url} 
\usepackage{subcaption}
\usepackage[font=small,labelfont=bf]{caption}
\usepackage{threeparttable}

\usepackage[T1]{fontenc}

\def\BibTeX{{\rm B\kern-.05em{\sc i\kern-.025em b}\kern-.08em
    T\kern-.1667em\lower.7ex\hbox{E}\kern-.125emX}}

\begin{document}

\title{Visual Marker Search for Autonomous Drone Landing in Diverse Urban Environments}

\author{
    \IEEEauthorblockN{Jiaohong Yao, Linfeng Liang, Yao Deng, Xi Zheng, Richard Han, Yuankai Qi}
    \IEEEauthorblockA{
        \textit{School of Computing, Macquarie University} \\
        Sydney, Australia \\
        jiaohong.yao@hdr.mq.edu.au
    }
}












\maketitle

\begin{abstract}

Marker-based landing is widely used in drone delivery and return-to-base systems for its simplicity and reliability. However, most approaches assume idealized landing site visibility and sensor performance, limiting robustness in complex urban settings. 
We present a simulation-based evaluation suite on the AirSim platform with systematically varied urban layouts, lighting, and weather to replicate realistic operational diversity. 
Using onboard camera sensors—RGB for marker detection and depth for obstacle avoidance—we benchmark two heuristic coverage patterns and a reinforcement learning–based agent, analyzing how exploration strategy and scene complexity affect success rate, path efficiency, and robustness. 
Results underscore the need to evaluate marker-based autonomous landing under diverse, sensor-relevant conditions to guide the development of reliable aerial navigation systems.

\end{abstract}


\section{Introduction}

\begin{figure}[htbp]
    \centering

    \includegraphics[width=1.0\linewidth]{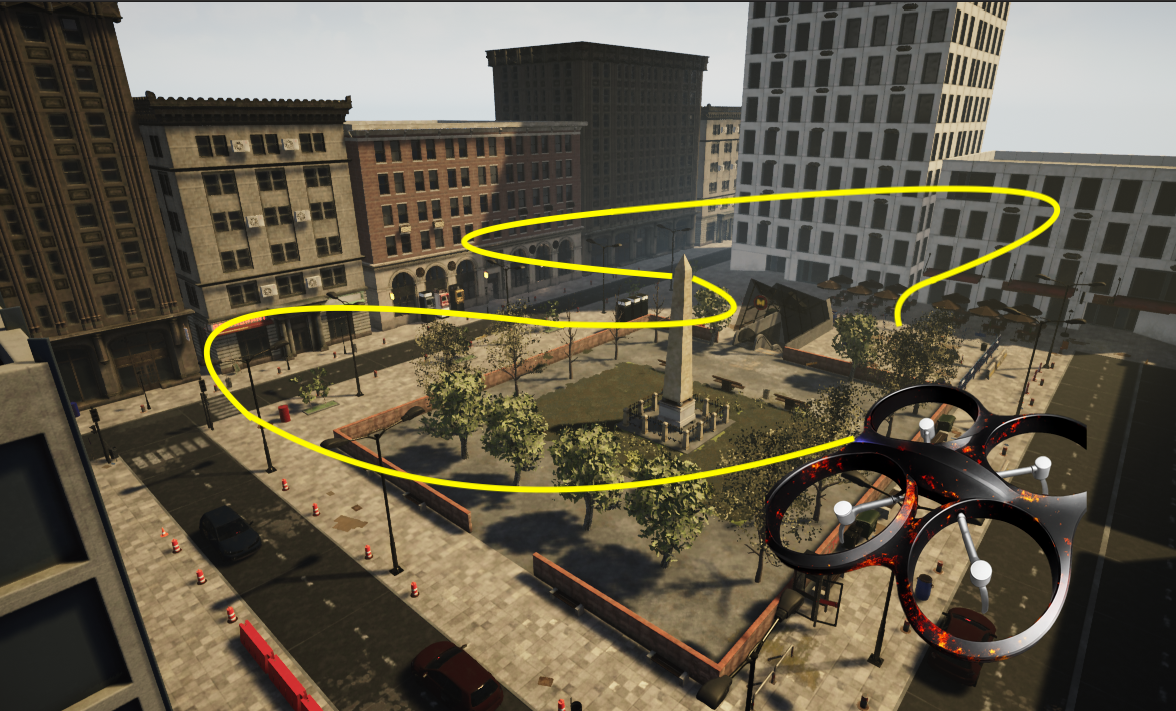}
    \caption*{Task Overview}

    \vspace{0.5em}

    \begin{subfigure}{0.3\linewidth}
        \centering
        \includegraphics[width=\linewidth]{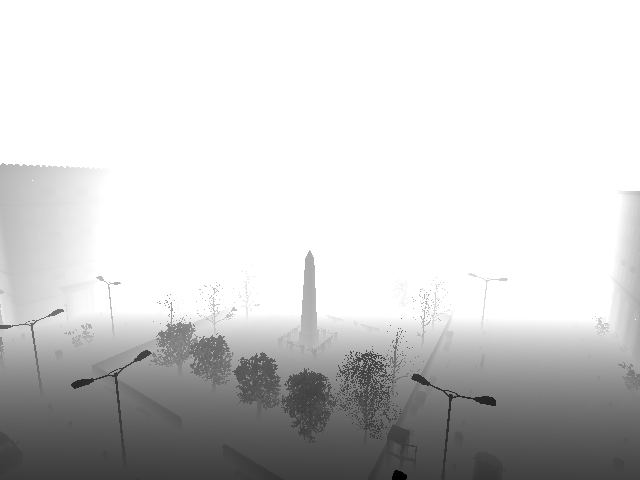}
        \caption*{Forward Depth}
    \end{subfigure}
    \hfill
    \begin{subfigure}{0.3\linewidth}
        \centering
        \includegraphics[width=\linewidth]{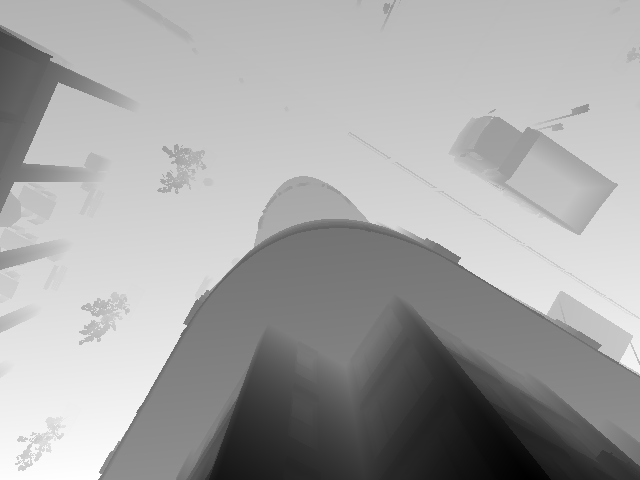}
        \caption*{Downward Depth}
    \end{subfigure}
    \hfill
    \begin{subfigure}{0.3\linewidth}
        \centering
        \includegraphics[width=\linewidth]{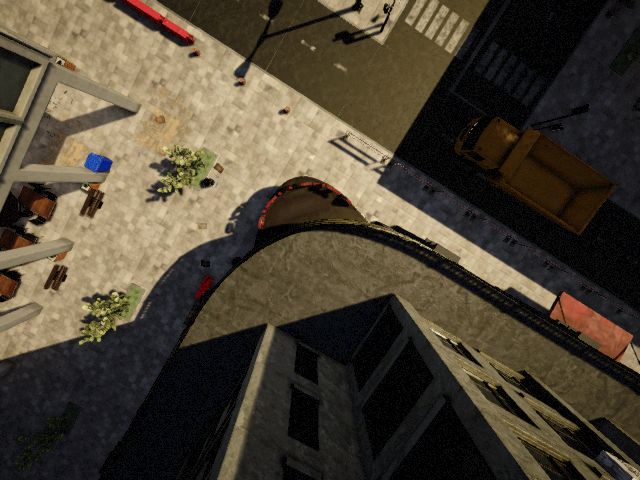}
        \caption*{Downward Scene}
    \end{subfigure}

    \vspace{0.5em}

    \caption{\textbf{Task description:} The agent autonomously searches for the marker within a fixed range. The yellow line illustrates an example trajectory, while the three images show the agent’s observations: depth for obstacle avoidance and RGB for marker detection.}
    \label{fig:overview_drone_obs}
\end{figure}



Visual-marker-based landing offers a practical solution for autonomous drone return and delivery, combining operational simplicity with reliable pose estimation at low cost~\cite{skyy, meituan_delivery}. By detecting visual markers at predefined landing zones using onboard cameras, drones can execute precise autonomous landings without relying on external infrastructure~\cite{schroder2025towards}.

Marker-based landing has demonstrated effectiveness in structured sites with reliable GPS, where global positioning guides drones to a predefined marker and visual localization finalizes descent~\cite{schroder2025towards}. Extending this capability to arbitrary urban locations—such as residential
rooftops or courtyards— is far more challenging, as GPS signals often degrade due to occlusion, multipath reflection, and weather-related interference. While sensors such as IMUs and barometers can partially enhance pose estimation, they cannot resolve GPS drift, leading to reduced  Extended Kalman Filter (EKF) accuracy, mapping inconsistencies, and unreliable navigation. Thus, robust landing in urban contexts requires active visual search and autonomous exploration guided by onboard sensors, substantially increasing task complexity.

\begin{figure}
    \centering
    \includegraphics[width=1.0\linewidth]{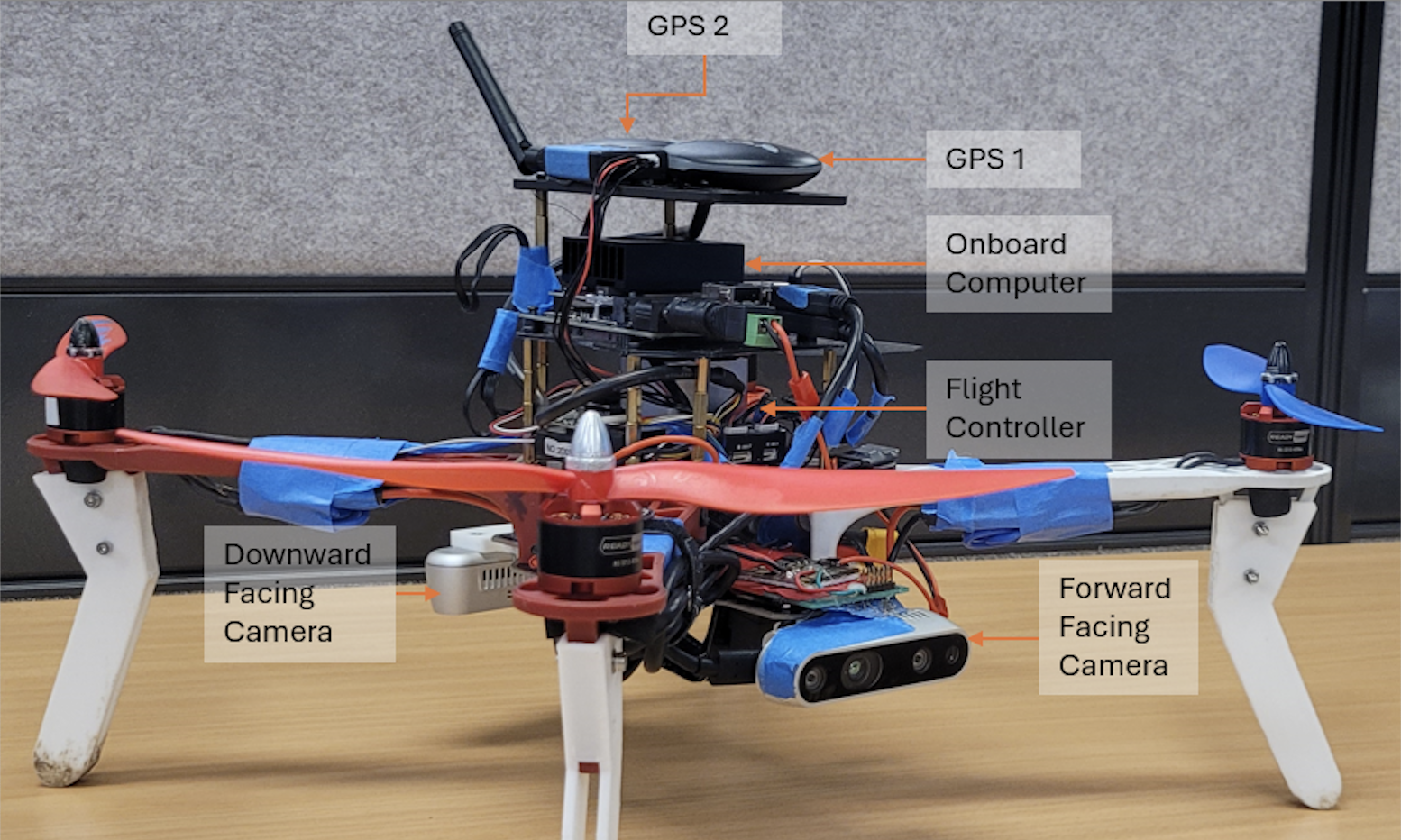}
    \caption{Drone System}
    \label{fig:drone_system}
\end{figure}

In urban environments, marker-based landing must handle variability in lighting, textures, structural layouts, and marker placement~\cite{schroder2025towards, liang2025garl}, which compromise perception and navigation given the sensing and computational limits of onboard hardware, as illustrated in Figure~\ref{fig:drone_system}. Lightweight platforms such as the Jetson Nano provide limited memory and processing, constraining real-time visual inference. While LiDAR, radar, and thermal sensors improve robustness under poor visibility, their cost, weight, and power demands hinder scalable deployment. In contrast, RGB and depth cameras offer a compact, low-power alternative aligned with practical constraints. Robust evaluation therefore requires datasets capturing diverse environments and realistic drone–environment interactions to systematically assess their influence on search and landing performance.

To address these challenges, this work emphasizes the importance of constructing diverse, interactive simulation datasets to improve system robustness and generalization. Rather than relying solely on traditional synthetic data augmentation~\cite{dadboud2025drift} or generating images for detection ability improvement~\cite{raxit2024yolotag, junior2023comparison, lee2024use}, we leverage the Unreal Engine 4~\cite{UE4} and Microsoft AirSim platform~\cite{airsim}, a high-fidelity simulator that facilitates dynamic interaction between drones and simulated environments. We manually select a wide range of urban landing sites with varying geometry, texture, contextual features, and environmental conditions—including diverse time-of-day and weather scenarios. At each simulated location, we strategically place markers and define corresponding drone initial positions, enabling drones to actively explore and locate landing targets.

    
    

Using this curated dataset, we evaluate three navigation strategies for autonomous marker search and landing. Experiments reveal clear differences in robustness and generalization, underscoring the impact of scenario diversity and environmental dynamics on system reliability. In summary, the main contributions are as follows:
\begin{itemize}
    \item A simulation-based evaluation suite with systematically varied scene attributes, including urban layouts, lighting, and weather.
    \item Benchmarks three navigation strategies across heterogeneous scenarios, providing insights into the relationship between exploration strategy, scene complexity, and task performance.
\end{itemize}


\section{Related Work}

Visual-marker-based landing approaches have become prevalent in drone delivery systems due to their cost-effectiveness and high localization precision~\cite{springer2024precision}.  Commercial systems, such as those used by Amazon Prime Air~\cite{wired2023primeair} and Zipline~\cite{zipline}, employ GPS to approach predefined marker-equipped sites, where visual algorithms guide the final descent~\cite{springer2024precision}. This paradigm, however, restricts deployment to known locations with reliable GPS~\cite{schroder2025towards}. To address this limitation, we propose a framework in which drones autonomously search for visual markers in previously unseen environments, enhancing adaptability and operational reach.

In the academic context, considerable effort has enhanced marker detection accuracy and robustness under varying visual conditions. For instance, Lee~\cite{lee2024use} has optimized marker designs (e.g., embedded ArUco patterns) to extend detection range and improve pose estimation accuracy. Deep learning has also been applied; Truong et al.~\cite{truong2020slimdeblurgan} combined motion-deblurring with object detection to improve performance under blur and low light. However, these studies typically assume static markers placed in controlled, simplified environments, largely overlooking complex and dynamic real-world conditions. Furthermore, they tend to approach marker search implicitly as a straightforward detection task, without addressing the broader, system-level challenge of autonomously locating markers. To bridge these gaps, we introduce an interactive, richly varied simulation dataset, explicitly modeling dynamic drone-environment interactions, aligning research with the practical complexities of real-world drone delivery tasks.


\section{Task}

As shown in Figure~\ref{fig:overview_drone_obs}, this task considers the problem of visually locating a small landing marker within a predefined circular region, without relying on GPS or prior map information. The drone is initialized at a random 6-DoF pose $P = [x, y, z, \text{yaw}, \text{pitch}, \text{roll}]$, while the marker is placed at an unknown
position within the environment. At each timestep $t$, the agent receives onboard observations $V_t = \{ I_t^D,\; D_t^D,\; D_t^F \}$, where $I_t^D$ and $D_t^D$ denote the downward-facing RGB and depth images used for marker detection, and $D_t^F$ is a forward-facing depth image used for obstacle avoidance. The agent follows a fixed flying pattern and altitude adjustment with a step size of 5 meters, or selects actions from a discrete set, including forward movement with a step size of 2 meters and rotation of 90 degrees. A navigation episode is regarded as successful if the estimated marker position lies within 2 meters of its ground-truth location. This formulation reflects realistic drone deployment scenarios and introduces challenges such as partial observability, limited visual context, environmental variability, and the small size of the marker for detection. 

\section{Dataset}

\subsection{Dataset Description}

\begin{figure}[htbp]
    \centering

    \begin{subfigure}{0.48\linewidth}
        \centering
        \includegraphics[width=\linewidth]{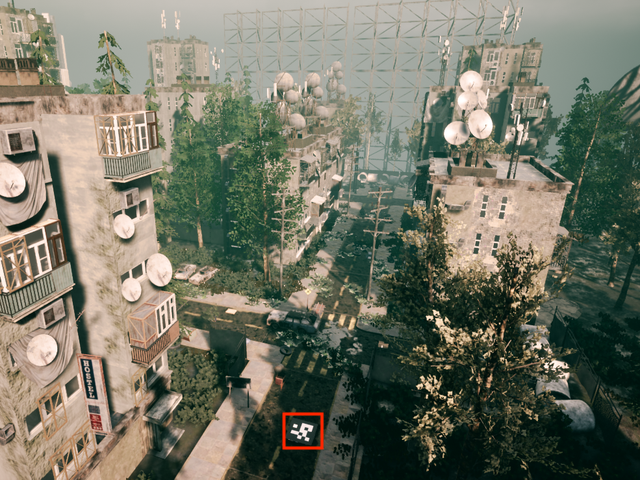}
        \caption*{PostSoviet sample 1}
    \end{subfigure}
    \hfill
    \begin{subfigure}{0.48\linewidth}
        \centering
        \includegraphics[width=\linewidth]{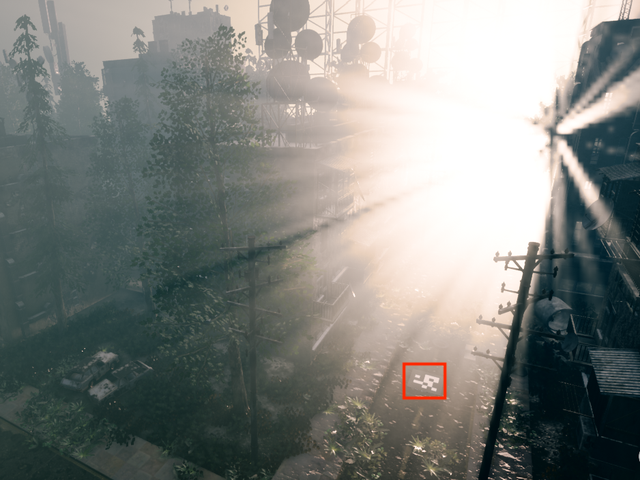}
        \caption*{PostSoviet sample 2}
    \end{subfigure}

    \vspace{0.5em}

    \begin{subfigure}{0.48\linewidth}
        \centering
        \includegraphics[width=\linewidth]{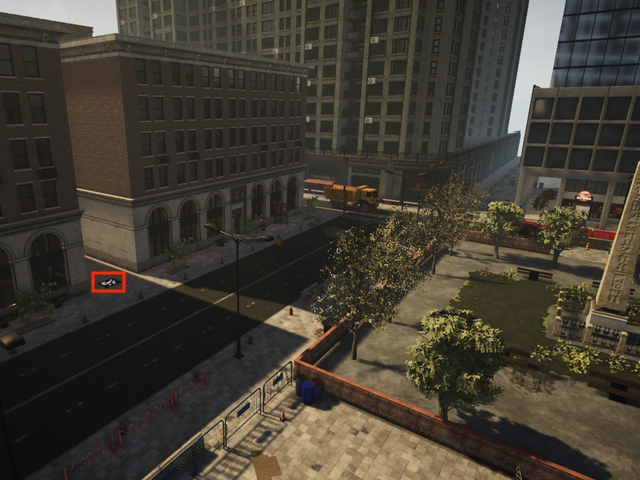}
        \caption*{ModernCity sample 1}
    \end{subfigure}
    \hfill
    \begin{subfigure}{0.48\linewidth}
        \centering
        \includegraphics[width=\linewidth]{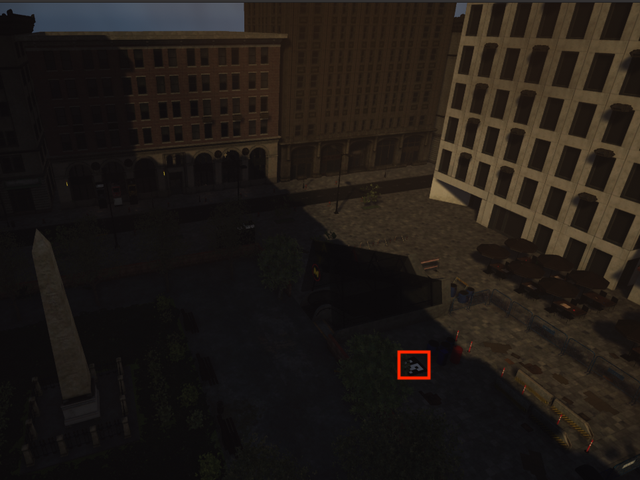}
        \caption*{ModernCity sample 2}
    \end{subfigure}

    \vspace{0.5em}

    \begin{subfigure}{0.48\linewidth}
        \centering
        \includegraphics[width=\linewidth]{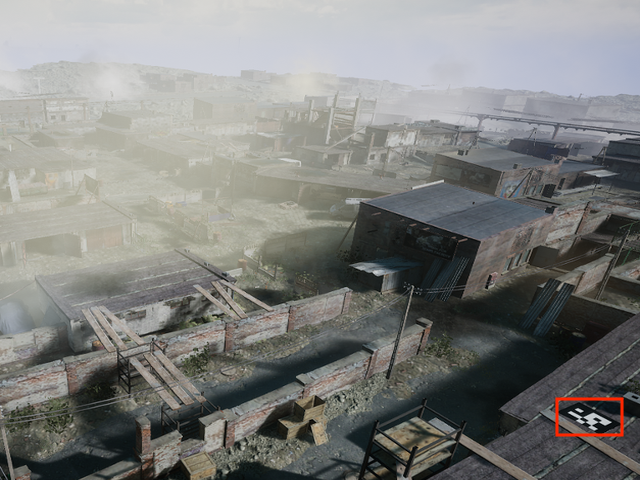}
        \caption*{UrbanDistrict sample 1}
    \end{subfigure}
    \hfill
    \begin{subfigure}{0.48\linewidth}
        \centering
        \includegraphics[width=\linewidth]{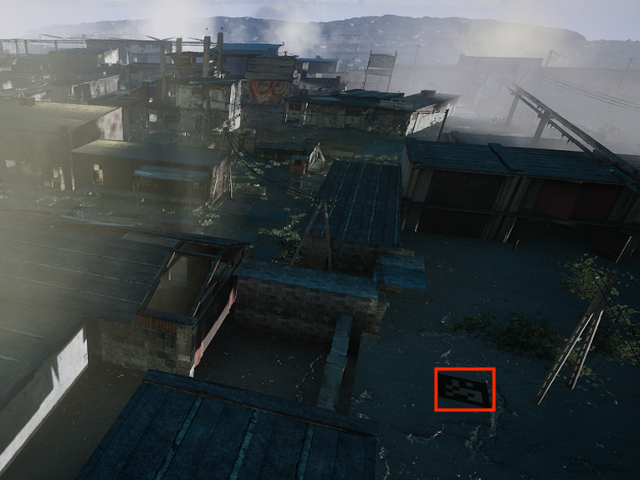}
        \caption*{UrbanDistrict sample 2}
    \end{subfigure}

    \caption{Examples of Marker Placement Across Environments (Markers Highlighted in Red)}
    \label{fig:example}
\end{figure}

To evaluate the generalization capabilities of vision-based marker search and landing methods, we construct a simulation-based dataset using Microsoft AirSim~\cite{airsim}, a high-fidelity platform built on Unreal Engine 4~\cite{UE4} that supports realistic urban environments, dynamic lighting, weather variation, and interactive drone control—factors critical for robust visual perception.

The dataset is built using three visually and structurally diverse urban maps~\cite{liu2023aerialvln} — ModernCity, PostSoviet, and UrbanDistrict — chosen to reflect diverse real-world delivery scenarios. ModernCity features modern residential buildings, open plazas, and landscaped areas. PostSoviet contains tall buildings interspersed with vegetation, simulating environments with dense vertical structures or natural obstructions. UrbanDistrict consists of low-rise, irregularly arranged buildings and narrow passages, capturing the visual and spatial complexity of compact city regions. Together, these environments provide diverse layouts, textures, and contextual variability for evaluating model generalization.

Within each environment, we manually construct marker–drone configurations to simulate realistic and non-trivial search tasks. Markers are placed at physically safe landing locations with diverse surrounding contexts, including variations in nearby objects, surface textures, and occlusions. Drone initializations are selected to avoid direct visibility of the marker, ensuring that successful detection requires active exploration rather than incidental exposure. To introduce perceptual diversity, we vary the time of day and weather conditions across scenarios, affecting global illumination, contrast, and visibility through changes in lighting, fog, and dust. In total, we generate 966 episodes across three environments: 102 in \textit{ModernCity}, 240 in \textit{PostSoviet}, and 624 in \textit{UrbanDistrict}. These are derived from 161 unique marker–drone combinations—comprising 38 distinct marker locations with up to five unique drone initializations each—augmented by variations in time of day, weather type, and severity to produce multiple scenario variants. This design captures diverse spatial and perceptual conditions representative of real-world urban navigation and landing challenges. Figure~\ref{fig:example} shows examples of marker placements across diverse urban environments under varied lighting (daytime, nighttime, glare, shadow) and visibility (fog, dust, occlusion) conditions. Markers are situated in cluttered, object-rich scenes, highlighting the dataset’s realism and its suitability for evaluating robust marker-based navigation.


Unlike prior datasets~\cite{truong2019deep, nguyen2018lightdenseyolo, kyristsis2016towards} that focus on static, easily visible landing sites for detection, our scenarios incorporate varied scene geometry, diverse visual contexts, and dynamic drone–environment interactions, creating realistic search complexity. This diversity enables robust evaluation of navigation strategies and systematic analysis of environmental factors (Section~\ref{sec:experiment}).

\subsection{Dataset Analysis}



\begin{figure}[htbp]
    \centering

    \begin{subfigure}{0.48\linewidth}
        \centering
        \includegraphics[width=\linewidth]{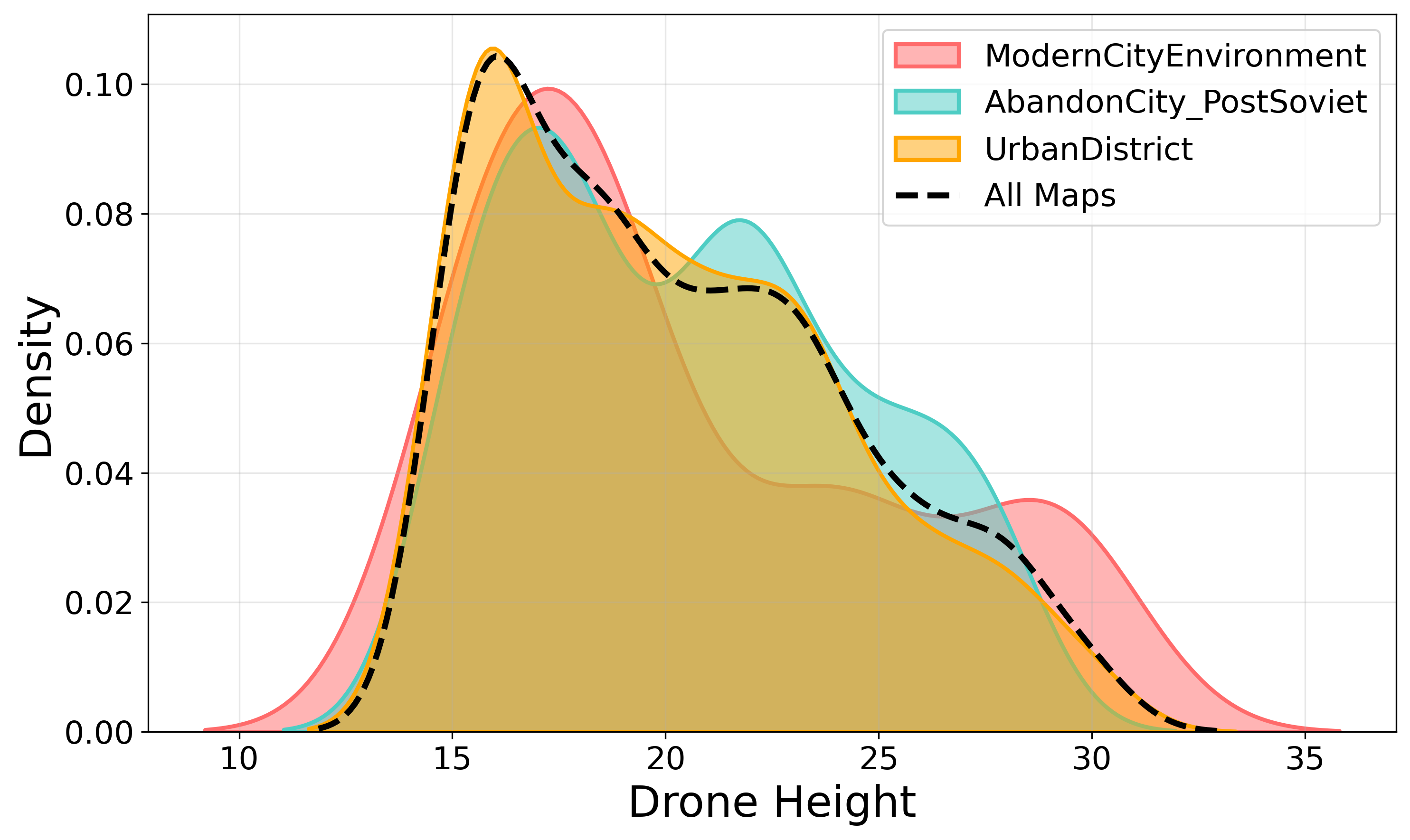}
        \caption*{Drone Height Distribution}
    \end{subfigure}
    \hfill
    \begin{subfigure}{0.48\linewidth}
        \centering
        \includegraphics[width=\linewidth]{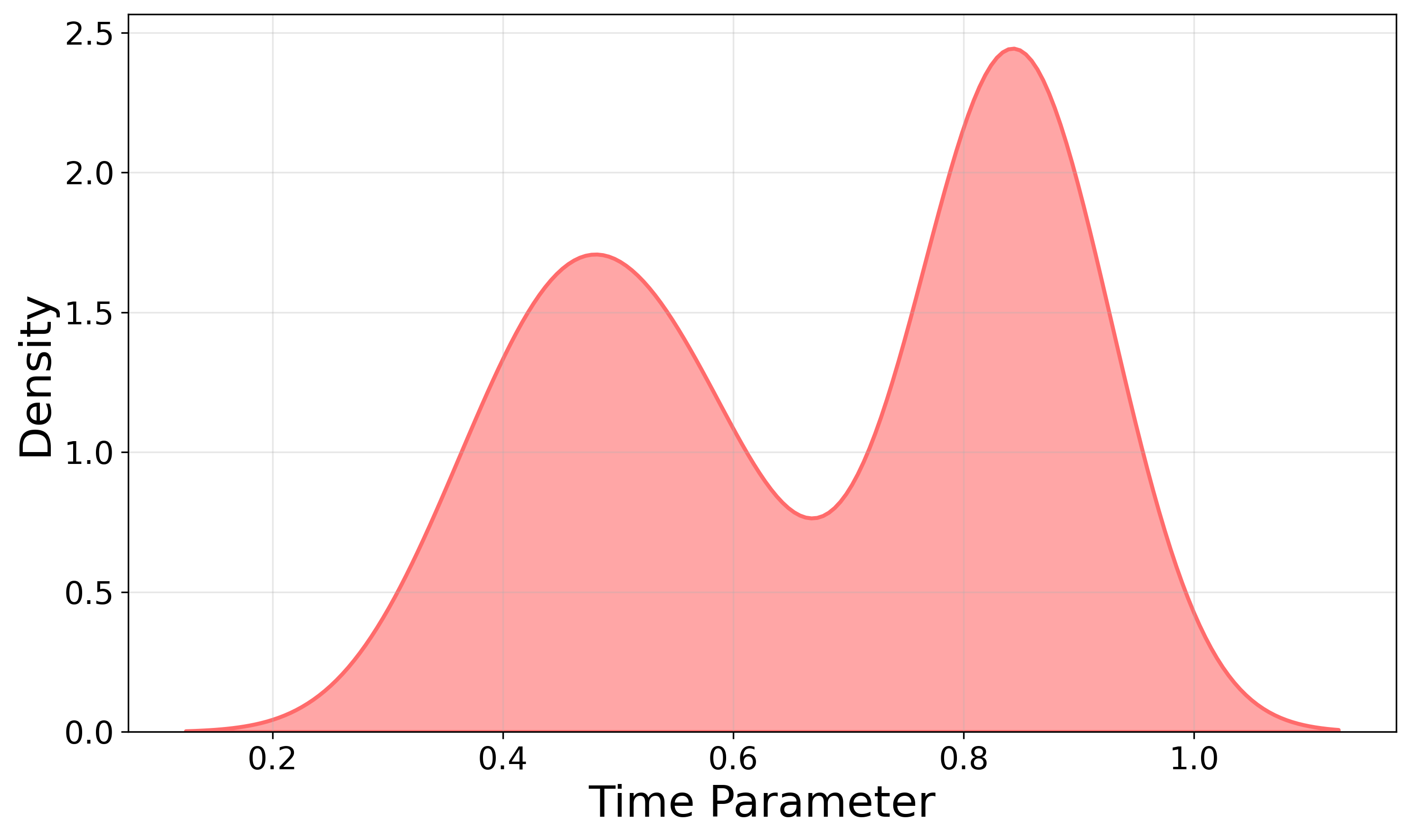}
        \caption*{Time Distribution}
    \end{subfigure}

    \vspace{0.5em}

    \begin{subfigure}{0.48\linewidth}
        \centering
        \includegraphics[width=\linewidth]{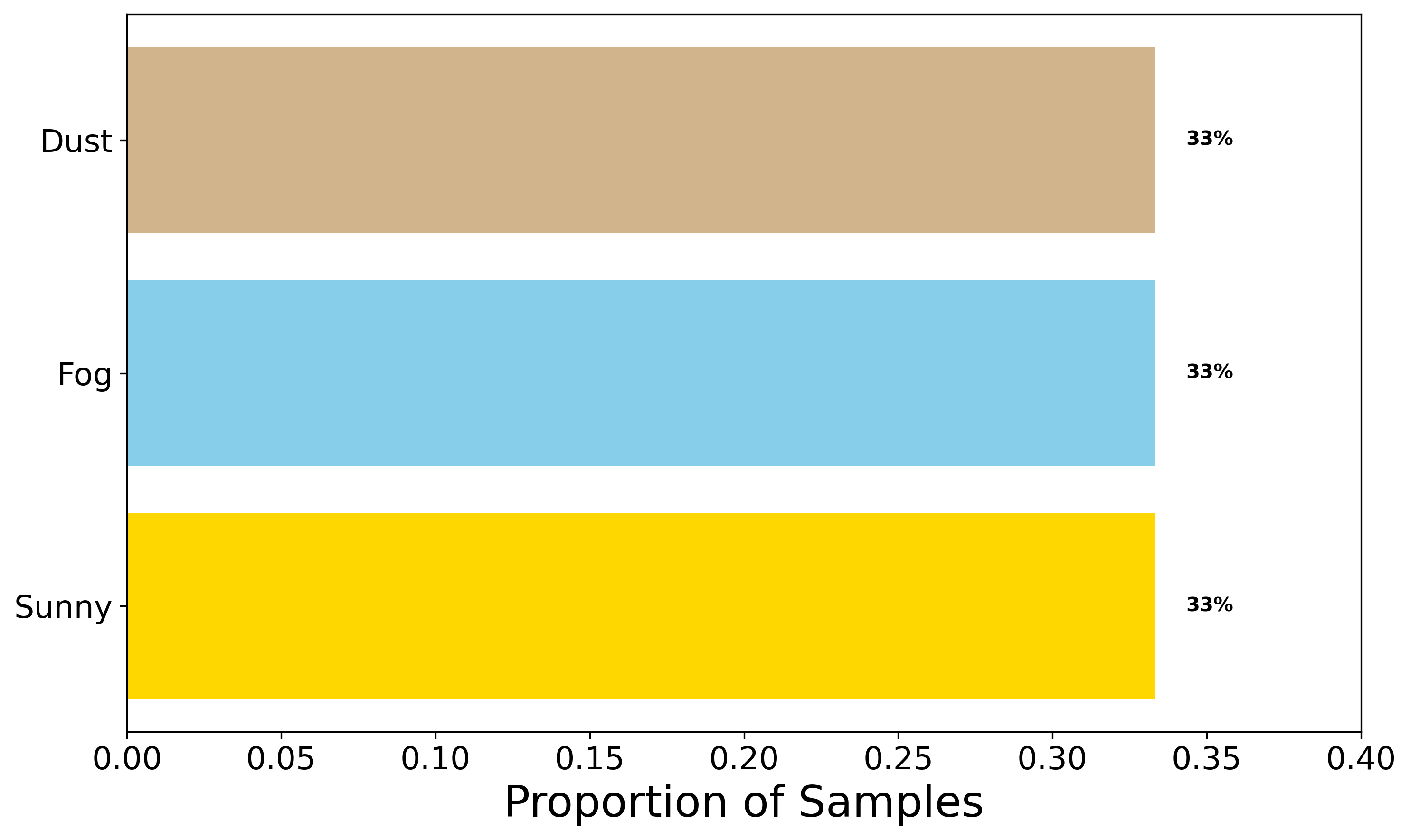}
        \caption*{Weather Type Distribution}
    \end{subfigure}
    \hfill
    \begin{subfigure}{0.48\linewidth}
        \centering
        \includegraphics[width=\linewidth]{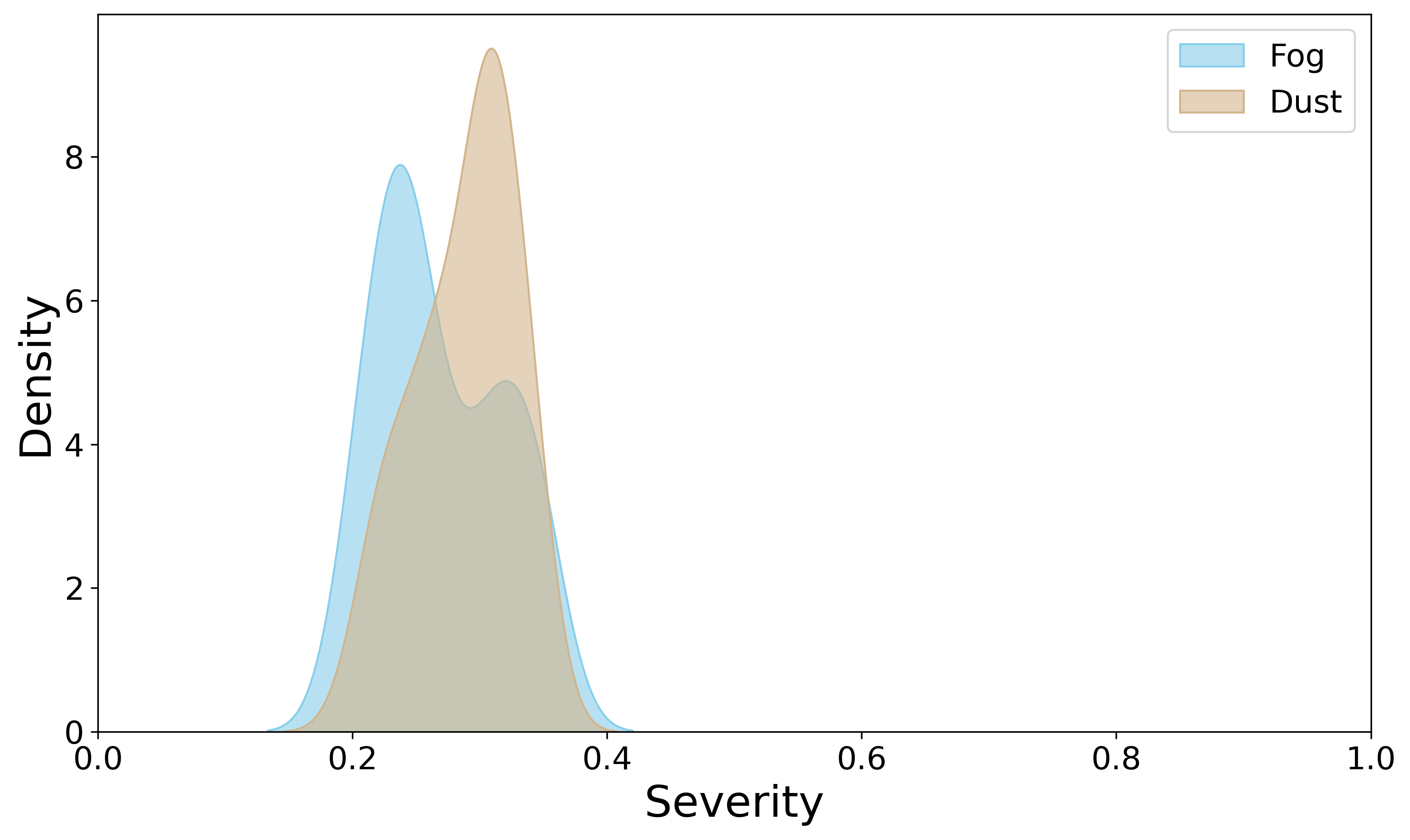}
        \caption*{Weather Severity Distribution}
    \end{subfigure}


    \caption{Parameter Distribution}
    \label{fig:param_distribution}
\end{figure}

The three simulated environments vary in spatial scale and layout—ModernCity (54 × 49 m), PostSoviet (59 m × 63 m), and UrbanDistrict (137 × 108 m), introducing differences in openness, openness, and structural density that shape the search dynamics. Figure~\ref{fig:param_distribution} summarizes the distribution of key scenario parameters contributing to environmental diversity. Drone heights range from 10 to over 30m, with most concentrated between 15–25m, reflecting structural differences across maps and dataset diversity. The time-of-day distribution is bimodal (peaks near 0.5 and 0.85), covering a range from midday to late afternoon. Weather types are evenly distributed among sunny, foggy, and dusty scenes ($\sim 33\%$ each), while severity levels are moderate (0.2–0.4), introducing perceptual challenges without hindering flight or detection. These variations provide a controlled yet diverse evaluation space for sensor-based navigation.
\section{Method}



We compare three navigation strategies: two heuristic baselines—Spiral~\cite{cao2025emergency} and Zigzag~\cite{liu2023path}—and a learning-based approach, E2E-RL~\cite{habitatchallenge2023}. All operate without external localization, relying solely on onboard sensors. Navigation terminates either upon successful marker detection in the downward-facing RGB image by a pretrained detector—used only at evaluation to keep policy learning independent of detector-specific biases—or when a method-specific stopping rule is met: for Spiral and Zigzag, completion of the predefined trajectory or a collision; for E2E-RL, exhausting the step budget, colliding, violating the search boundary, or exhibiting sustained non-progressive behavior.


\textbf{Spiral}~\cite{cao2025emergency} follows a predefined outward-expanding pattern centered at the drone’s initial position, incrementally sweeping a circular region in the horizontal plane. Both 2D and 3D variants share the same fixed step-size trajectory at constant altitude; the 3D version additionally ascends only when necessary to avoid obstacles and returns to the original height afterward.

\textbf{Zigzag}~\cite{liu2023path} traverses the environment using a structured back-and-forth sweep along one axis, interleaved with periodic shifts along the perpendicular axis. Like Spiral, waypoints are computed in advance with fixed spatial intervals, and 3D variates can adjust altitude temporarily in response to obstacles.

\textbf{E2E-RL}, which follows the design principles of ground object-goal navigation agents~\cite{habitatchallenge2023}, is adapted for aerial visual marker search in complex urban environments. The agent learns an adaptive exploration policy from real-time onboard observations, receiving each timestep a front-view depth image and relative positional input, and selecting from a discrete 2D action set (move forward, turn left/right). Depth observations encode geometric structure, obstacle size and shape, relative distances, and spatial relationships—enabling the policy to infer promising directions for maximizing coverage and avoiding obstacles. The policy is optimized via reinforcement learning with a reward function that encourages coverage efficiency and penalizes collisions and unnecessary rotations, supporting perception-driven navigation without predefined trajectories or maps.


\section{Experiment}
\label{sec:experiment}

\subsection{Experiment Setup}
We evaluate these navigation strategies over 966 simulated episodes with varied drone initializations, marker placements, lighting, and weather. The search area is a circle with a radius of 30 meters,
centered at the drone’s start position. All methods use identical sensor configurations, with 640$\times$480 depth images for obstacle avoidance and downward-facing RGB images for marker detection. During evaluation, marker detection is performed by a YOLOv11~\cite{khanam2024yolov11} model fine-tuned on domain-specific aerial imagery under varied altitudes and conditions, using a confidence threshold of 0.8. The detector is excluded from training to prevent the E2E-RL agent from exploiting detector-specific triggers instead of learning genuine exploration strategies. The E2E-RL policy is trained with PPO~\cite{schulman2017proximal} and curriculum learning~\cite{bengio2009curriculum} entirely in a lightweight surrogate environment that preserves key task properties—partial observability, occlusions, and 3D obstacle avoidance—using only 64$\times$64 front-view depth images to maximize coverage efficiency while avoiding collisions.

\subsection{Evalaution Metrics}
To evaluate navigation performance, we adopt five metrics commonly used in embodied visual navigation tasks~\cite{zhang20233d,xu2024aligning,dang2023search}. Success Rate (SR) measures the proportion of episodes in which the agent stops within 2 meters of the marker’s true 3D location. Navigation Error (NE) calculates the Euclidean distance between the agent’s final stopping position and the ground-truth marker position. Success weighted by Path Length (SPL) reflects trajectory efficiency by weighting SR with the ratio of the shortest possible path length to the actual path length taken by the drone. To evaluate additional aspects of robustness, we further report Collision Rate (CR), defined as the fraction of episodes where the agent collides with obstacles, and False Detection Rate (FD), which quantifies the fraction of episodes where the detector incorrectly signals a successful marker detection but the reported position lies more than 2 meters away from the actual marker location.


While metric definitions follow standard conventions, we introduce practical adjustments. For SR and NE, the agent’s final position is defined as the 3D location of the first detected marker, assuming reliable delivery via position control once detection occurs~\cite{schroder2025towards}. SPL and CR are instead computed over the exploration trajectory up to detection, with the shortest-path length in SPL approximated by the straight-line distance in the $xy$-plane to treat vertical movement consistently. This separation ensures fair comparison of navigation behavior without conflating it with post-detection execution.

\subsection{Results and Analysis}


\subsubsection{Model Performance Analysis}

\begin{table}[htbp]
\centering
\caption{Performance  across methods and maps}
\setlength{\tabcolsep}{0.5pt}
\begin{tabular}{llccccc}
\toprule
\textbf{Method} & \textbf{Map} 
& \textbf{SR/\%$\uparrow$} 
& \textbf{NE/m$\downarrow$} 
& \textbf{SPL/\%$\uparrow$} 
& \textbf{CR/\%$\downarrow$} 
& \textbf{FD/\%$\downarrow$} \\\midrule
\multirow{4}{*}{Spiral-2D~\cite{cao2025emergency}} 
& ModernCity     & 61.76 & 13.20 & 5.02 & 22.55 & 3.93 \\
& UrbanDistrict  & 60.90 & 17.41 & 4.73 & 0.00  & 8.18 \\
& PostSoviet     & 36.67 & 20.46 & 2.82 & 47.50 & 0.00 \\
& \textbf{Avg} & \textbf{54.97} & \textbf{17.72} & \textbf{4.29} & \textbf{14.18} & \textbf{5.70} \\
\midrule
\multirow{4}{*}{Zigzag-2D~\cite{liu2023path}} 
& ModernCity     & 37.25 & 22.34 & 2.13 & 47.06 & 3.93 \\
& UrbanDistrict  & 62.82 & 16.13 & 3.51 & 0.00  & 3.05 \\
& PostSoviet     & 32.92 & 21.74 & 1.90 & 50.00 & 0.00 \\
& \textbf{Avg} & \textbf{52.69} & \textbf{18.18} & \textbf{2.96} & \textbf{17.39} & \textbf{2.39} \\
\midrule
\multirow{4}{*}{Spiral-3D} 
& ModernCity     & 63.73 & 14.51 & 5.17 & 14.70 & 9.80 \\
& UrbanDistrict  & 60.90 & 17.41 & 4.73 & 0.00  & 8.18 \\
& PostSoviet     & 57.92 & 17.16 & 4.69 & 0.00  & 1.66 \\
& \textbf{Avg} & \textbf{60.63} & \textbf{17.03} & \textbf{4.79} & \textbf{2.37} & \textbf{6.80} \\
\midrule
\multirow{4}{*}{Zigzag-3D} 
& ModernCity     & 57.84 & 16.76 & 3.35 & 5.88 & 3.92 \\
& UrbanDistrict  & 62.82 & 16.13 & 3.51 & 0.00 & 3.05 \\
& PostSoviet     & 47.92 & 20.82 & 2.92 & 0.00 & 0.83 \\
& \textbf{Avg} & \textbf{58.66} & \textbf{17.38} & \textbf{3.34} & \textbf{1.23} & \textbf{2.87} \\
\midrule
\multirow{4}{*}{E2E-RL~\cite{habitatchallenge2023}} 
& ModernCity     & 18.63 & 30.51 & 9.87 & 0.00 & 7.84 \\
& UrbanDistrict  & 30.13 & 26.37 & 19.22 & 0.00 & 5.61 \\
& PostSoviet     & 17.08 & 31.66 & 11.95 & 0.00 & 9.17 \\
& \textbf{Avg} & \textbf{24.38} & \textbf{28.27} & \textbf{15.49} & \textbf{0.00} & \textbf{6.53} \\
\bottomrule
\end{tabular}
\label{tab:method-map-performance}
\end{table}

Table~\ref{tab:method-map-performance} compares the five navigation strategies across multiple metrics. The heuristic baselines, Spiral-2D and Zigzag-2D show an acceptable success rate (SR) but high collision rates (CR), where the improved versions Spiral-3D and Zigzag-3D achieve a higher SR and lower navigation errors (NE) overall, benefiting from their predefined coverage patterns that ensure systematic spatial visibility. Spiral slightly outperforms Zigzag in SR, while Zigzag attains marginally lower false detection rates (FD), likely due to denser local coverage. In contrast, E2E-RL shows notably lower CR and higher SPL, reflecting more direct and efficient trajectories, which is valuable in real-world settings, where energy, time, and airspace constraints limit exhaustive search. This advantage arises from its use of real-time depth observations, which encode geometric structure, obstacle size and shape, relative distances, and spatial relationships, allowing the policy to infer promising directions for coverage and obstacle avoidance without predefined paths. 
Although its SR and NE are suboptimal, partly due to less exhaustive coverage and constrained vertical mobility, the consistently higher SPL and lowest CR suggest that E2E-RL exploits the environment intelligently, highlighting its potential as an adaptive alternative to deterministic strategies.


As shown in Table~\ref{tab:method-map-performance}, environmental layout significantly influences the effectiveness of each navigation strategy. In UrbanDistrict, the flat terrain and dense arrangement of low-rise, irregular buildings result in minimal vertical occlusion but increased horizontal clutter. This configuration favors coverage-based methods—Spiral and Zigzag—whose sweeping aligns well with surface-level visibility, yielding high SR and low NE. PostSoviet, by contrast, introduces greater vertical complexity, with tall structures and vegetation creating obstacles, occlusions and shadowed regions. These features impair line-of-sight and contribute to reduced SR and high CR. ModernCity offers a semi-structured layout, with mid-rise buildings enclosing an open courtyard. This predictable geometry supports consistent performance across methods, though Spiral’s uniform scanning may over-trigger detections in visually homogeneous regions, explaining its higher FD rate. Overall, the results highlight how obstacle density, spatial organization, and visual structure affect marker search performance.

\subsubsection{Environmental Factors Analysis}

\begin{table}[htbp]
\centering
\caption{Environmental Correlates of Navigation Outcomes. This table summarizes average scene attributes associated with different navigation outcomes (Success, Failure, or False Detection). \textit{Height (m)} refers to the drone’s initial flight altitude. \textit{Time} denotes the normalized time-of-day parameter. \textit{Sunny (\%)} represents the proportion of episodes conducted under clear weather conditions. \textit{Severity (\%)} captures the intensity of visual degradation in non-sunny environments (i.e., fog or dust).}

\setlength{\tabcolsep}{3pt}
\begin{tabular}{llcccc}
\toprule
\textbf{Method} & \textbf{Type} 
& \textbf{Height/m} 
& \textbf{Time} 
& \textbf{Sunny/\%} 
& \textbf{Severity/\%} \\
\midrule
\multirow{3}{*}{Spiral-3D}
& Success         & 19.01 & 0.67   & 33.00 & 18.53 \\
& Fail            & 22.82 & 0.69   & 33.00 & 18.44 \\
& FD & 18.72 & 0.67   & 38.00 & 16.65 \\
\midrule
\multirow{3}{*}{Zigzag-3D}
& Success         & 18.76 & 0.68   & 34.00 & 18.13 \\
& Fail            & 22.60 & 0.68   & 33.00 & 18.65 \\
& FD & 19.24 & 0.80   & 28.00 & 19.98 \\
\midrule
\multirow{3}{*}{E2E-RL~\cite{habitatchallenge2023}}
& Success         & 18.77 & 0.67 & 35.08 & 17.83 \\
& Fail            & 20.86 & 0.69 & 33.08 & 18.50 \\
& FD & 19.89 & 0.65 & 29.23 & 19.25 \\
\bottomrule
\end{tabular}
\label{tab:scene-attr-by-method}
\end{table}

Table~\ref{tab:scene-attr-by-method} highlights how spatial and perceptual factors such as altitude, lighting, and visibility influence marker detection stability and overall search performance.

\textbf{Drone height} strongly correlates with navigation success. For all methods, failed episodes occur at significantly higher initial altitudes—exceeding 20.8m for E2E-RL and 22.6m for the heuristic baselines—likely due to diminished marker size and reduced marker visibility at elevated viewpoints. In contrast, both successful and false detection (FD) cases concentrate around lower altitudes (18.7–19.9m), suggesting that FD errors are not simply due to limited visibility, but rather arise even when the marker is within view.

\textbf{Time of day}, approximated by the normalized time parameter, is largely consistent across across success, failure, and false detection episodes for most methods, indicating that general lighting conditions are not a primary factor affecting success or false detection. A notable exception is Zigzag’s FD, which show a higher average time value—potentially reflecting more challenging illumination such as dark or shadow.

\textbf{Weather conditions}, measured by visibility severity and sunny ratio, show modest but method-dependent effects. For Zigzag and E2E-RL, FD are more prevalent under lower visibility (i.e., higher severity, reduced sunny ratio), suggesting increased perceptual ambiguity in degraded scenes. In contrast, Spiral exhibits more false positives under clearer conditions, potentially due to its uniform coverage and frequent angle changes, encountering sunlit textures that resemble the marker. This contrast highlights divergent failure modes: while learning-based methods are more vulnerable to visual degradation, coverage-based approaches may over-trigger under high-visibility settings due to less selective filtering.

\section{Conclusion}



In this work, we develop a new simulated platform for evaluating visual marker search, navigation, and landing strategies under realistic urban conditions. Through systematic variation of layout, lighting, and weather, we identify key factors affecting method performance and generalization. This platform offers a controlled testbed for advancing robust aerial autonomy, with future work targeting improved policy adaptability via 3D exploration and richer visual cues such as RGB appearance.

\bibliographystyle{IEEEtran}
\bibliography{IEEEabrv,sections/reference}

\end{document}